# AI-Driven Robotic Crystal Explorer for Rapid Polymorph Identification

Edward C Lee[†], Daniel Salley[†], Abhishek Sharma and Leroy Cronin*

[1]School of Chemistry, University of Glasgow, University Avenue, Glasgow G12 8QQ, UK.

*Corresponding author Email: Lee.Cronin@glasgow.ac.uk   [†]These authors contributed equally.

**Crystallisation is an important phenomenon which facilitates the purification as well as structural and bulk phase material characterisation using crystallographic methods. However, different conditions can lead to a vast set of different crystal structure polymorphs and these often exhibit different physical properties, allowing materials to be tailored to specific purposes. This means the high dimensionality that can result from variations in the conditions which affect crystallisation, and the interaction between them, means that exhaustive exploration is difficult, time-consuming, and costly to explore. Herein we present a robotic crystal search engine for the automated and efficient high-throughput approach to the exploration of crystallisation conditions. The system comprises a closed-loop computer crystal-vision system that uses machine learning to both identify crystals and classify their identity in a multiplexed robotic platform. By exploring the formation of a well-known polymorph, we were able to show how a robotic system could be used to efficiently search experimental space as a function of relative polymorph amount and efficiently create a high dimensionality phase diagram with minimal experimental budget and without expensive analytical techniques such as crystallography. In this way, we identify the set of polymorphs possible within a set of experimental conditions, as well as the optimal values of these conditions to grow each polymorph.**

Crystal polymorphism occurs when a compound can form multiple distinct crystal structures.[1] Polymorphs exhibit different physical[2], spectroscopic[3], surface[4], mechanical[5] and chemical properties[6], and as such, identification and reliable separation of different crystal polymorphs is vital in many fields[7]. Additionally, for active pharmaceutical ingredients, patenting laws protect the polymorph, not the molecule[3]. Polymorphism arises due to the effects of specific crystallisation conditions on inter- and intra-molecular interactions, such as hydrogen bonding, pi-stacking, and Van der Waals forces, which affect the molecular orientation of and between nucleating molecules[8–10]. This gives rise to three possible features which may or may not be desired: concomitant crystallisation of different polymorphs, preferential crystallisation of polymorphs with sup-optimal properties, and spontaneous conversion to a more thermodynamically stable polymorph[9].

For these reasons, a comprehensive knowledge of a compound's polymorphs and polymorph formation conditions is important for reproducibility, scaling, and yield optimisation. However, any exhaustive search to obtain all of a compound's polymorphs is both resource and time-expensive due to the high dimensionality of parameters that affect crystallization that must be explored. While high-throughput automation has helped, inefficient strategies such as Grid Search (GS)[11] are often applied with the aim of checking all conditions to a finite resolution, which is determined by available resources. GS is a sub-optimal strategy due to an excess of sampling points in regions where there is high outcome certainty, and a deficit of sampling points where outcome uncertainty is low, and this results in an inefficient use of any experimental budget. A better approach is to use an exploration/optimisation strategy combined with active learning, where sampling points are determined by a function operating on some measurable feature of the system being investigated, a method that has shown promise in many areas of chemistry [12–14]. However, applying this strategy to polymorph exploration presents two main challenges. Firstly, the only relevant observable feature is the relative yield of each polymorph, which is hard to quantify without expensive manual methods such as crystallography, and secondly, the response surfaces for polymorph yields are typically flat, making exploration and optimisation by yield alone impossible.

To resolve both of these problems, we have developed a method that uses an automated high-throughput closed-loop approach to automatically quantify relative polymorphic yields using computer vision, and optimise the search strategy to prioritise regions of crystallisation space with the highest uncertainty using Bayesian exploration, see Figure 1. In addition to being able to quantify the amount of each known polymorph in a sample, this approach is able to identify the presence of any new polymorph discovered, which may be characterised outside of the closed loop. Once characterised, the conditions required to form new discovery undergo Bayesian optimisation, whereby, the relative amount (as determined by computer vision) of the new polymorph is optimised. Here, we show how the combination of polymorph identification and quantification computer vision can be used together with a Bayesian exploration/optimization strategy can efficiently explore a crystallisation space comprised of the relative amounts of four solvents for the compound 5-methyl-2-[(2-nitrophenyl)amino]thiophene-3-carbonitrile, build a phase diagram and discover the conditions to form a rare polymorph in the crystallisation space.

**Platform:** The automated platform was inspired by previous designs[15] and consisted of a liquid handling and crystallisation robot with a capacity of 72 parallel crystallisations from a total solvent volume of 2.5 mL, as well as automated imaging using a camera positioned below the crystallisation vials. The platform itself and its significant components/assemblies can be seen in Figure 2 and details of its construction and operation are described in SI Section 2.

**Crystal-vision:** A computer-vision based library, named crystal-vision, was developed for automated detection, classification, and segmentation of *in situ* crystal images based on Mask

RCNN[16] computer vision by loading a pre-trained image classification model and retraining the final layers on images of crystals. This enabled the creation of image analysers which could be used to infer the presence, type, location, and size of crystals in a sample vial.

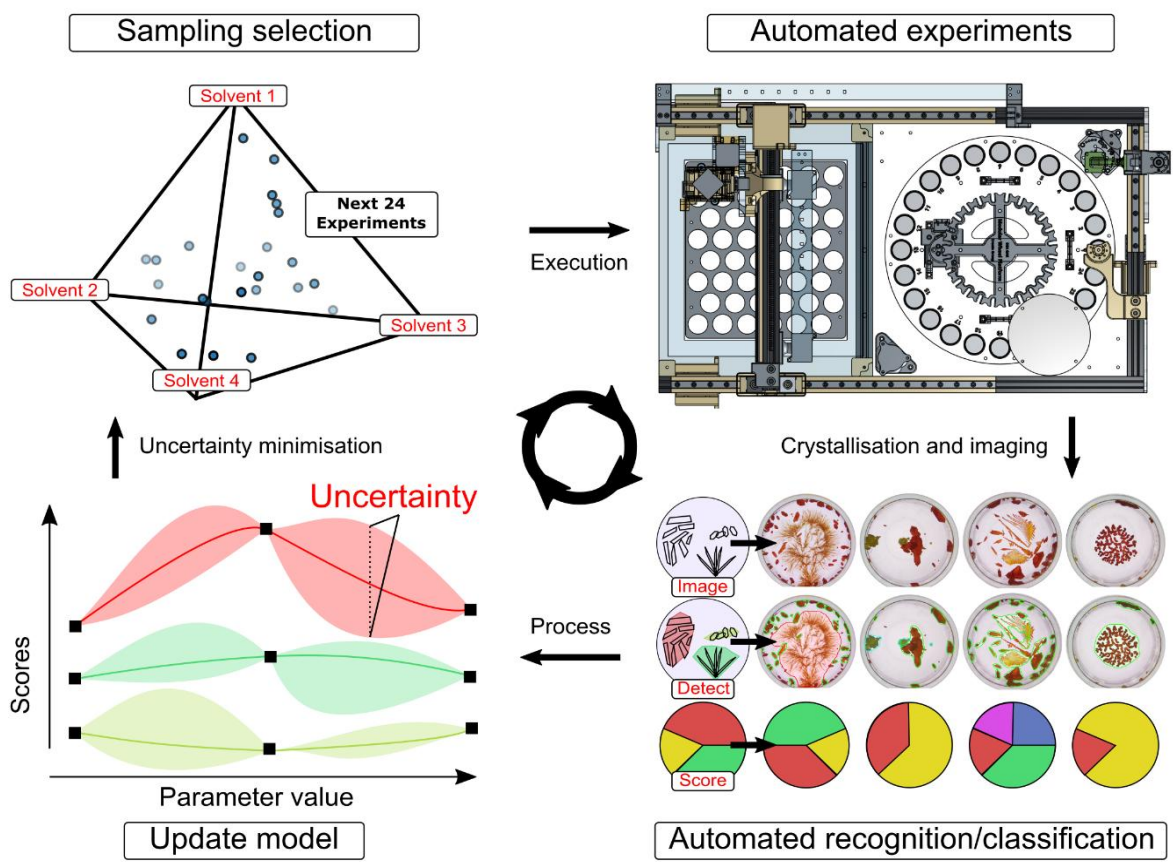

**Figure 1 Polymorph discovery loop**. Top left: Crystallisation conditions are chosen from a parameter space initially at random and subsequently based on previous findings. Top right: The automated platform then prepares the crystallisation solutions as specified. Bottom right: Experimental samples of crystallised material are imaged using a high-definition camera. Images are segmented into crystal/non-crystal using computer vision, as well as into different polymorph classes. New polymorphs are identified as the areas where the difference between the set of "crystal" pixels and the union of all "polymorph" pixels is greater than 0. Bottom Left: these data are then used to create a Bayesian surrogate model from which the points with the highest uncertainty are selected in subsequent generations.

The detection capability was used to distinguish between crystal presence and absence, whereas classification was used to identify what subclass of crystal had been identified. The classifier was trained with two subclasses: morphology and identity. The morphology classifier could distinguish between amorphous/non-crystalline material, powder crystalline samples, overlapping crystals/crystal clusters, single/isolated crystals, and particular features of crystals such as tracht. The identity classifier could identify the compound of the crystal, provided that it had been previously incorporated into the classifier training database. Training details are provided in SI 3.1 and 3.2

Together, these capabilities allowed the detection of crystallisation onset time, and crystalline quality of a specific compound or polymorph. With the inclusion of image segmentation, the size and location of each crystal could be detected, and thus growth rates could be monitored for multiple single crystals simultaneously in one reaction vial. Alternatively, the system could be used to identify the presence of a previously unobserved crystal. This method involved using a classifier which had been trained on many types of crystals (classifier A) to recognise whether something is or is not a crystal. Another classifier which had been trained to recognise the specific identity of a crystal (classifier B) could then be applied to attain the identity of crystals in the sample. A positive response in a region from classifier A but a negative response in the same region from classifier B implies that the region corresponds to a crystal whose identity is unknown, and therefore potentially a new discovery. Initially, classifier B would suffer from overfitting due to the lack of data at the start of the experiment. However, this problem could be mitigated by incorporating more data as they were collected (see SI Section 3).

**Exploration:** The strategy for polymorph search and discovery involved a Bayesian methodology, where data from previous experiments were used to construct a model of the polymorphic system from Gaussian Processes. The investigated parameters were the proportions of each solvent in the crystallising system and the observed parameters were the proportions of each polymorph detected by the crystal-vision detector. In this way, separate surrogate models could be constructed for each polymorph to predict the likelihood of each polymorph's formation at any point, as well as the uncertainty (variance) associated with this. The aim of exploration was to obtain as accurate a model as possible across all investigated parameters in as few experiments as possible, increasing the likelihood that evaluation points which result in novel or rare outcomes are located faster than by methods such as grid search.

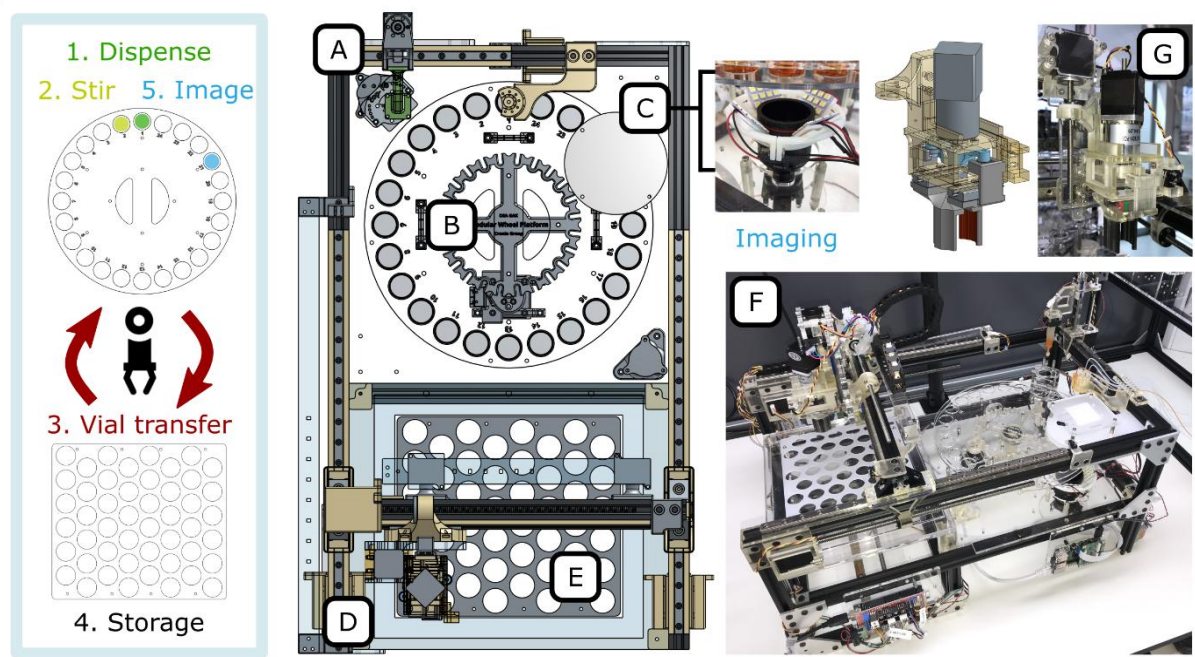

**Figure 2: Fully automated robotic platform.** Left side of the figure gives a brief overview of the significant steps the platform performs for each sample generation. (A) Shows the position of the overhead stirring assembly, (B) 24 vial sample wheel, (C) Imaging set-up with C-mount Raspberry Pi Camera lens on a HQ Raspberry Pi camera, (D) X and Y-axis motors for the gripping assembly, (E) vial Storage area, (F) Complete platform as built, (G) CAD and actual image of vial transfer gripper.

Specifically, crystallisation conditions over multiple generations were to maximise the Negative Integrated Posterior Variance of the priors of the surrogate model. This was undertaken in an automated semi-closed loop process using an automated platform and an active learning search algorithm using the BoTorch package[17] (See Figure 1). A set of simultaneous crystallisations were performed on the automated platform by mixing samples of the dissolved compound with other solvents at 25°C and 1 atm pressure, then waiting 12 hours for slow evaporation and crystallisation to occur. At this point, the vials were robotically transferred to an imaging platform, and the images were analysed using two classifiers created using the crystal-vision library to detect and identify known crystals and to alert the user to any unknown crystals, as described above. This process was simulated in silico (SI 3.3) with a random acquisition function being compared to NIPV.

Upon detection of an unknown crystal, the closed loop would be interrupted to allow for verification of the new polymorph by x-ray diffractometry. On confirmation of a new polymorph, the producing conditions were repeated multiple times in order to obtain a larger training dataset of the polymorph, and the classifier was retrained incorporating these.

Once scores for each polymorph in each reaction had been calculated, they were incorporated into a Bayesian Network and a surrogate model was built from the posterior probabilities using Gaussian Processes. Since we were interested in prioritising exploration of uncertain regions in the reaction space, subsequent reaction conditions were then chosen to minimise the uncertainty in the surrogate models of each polymorph, specifically by applying an algorithm which would choose a reaction that would minimise negative integrated posterior variance (NIPV). This allowed an efficient reaction selection routine to maximise exploration away from known/expected outcomes in a system with high input and output dimensionality. Once solvent mixtures were selected, the automated platform was able to perform a subsequent generation of crystallisations and continue the loop.

The compound 5-methyl-2-[(2-nitrophenyl)amino]thiophene-3-carbonitrile (also known as ROY) is one of the most polymorphic compounds known (12 forms reported to date)[18] was selected as a potential candidate for the discovery of further polymorphic forms. It also has the property that several of its differently coloured polymorphs can form under similar conditions, resulting in concurrent polymorph crystallisation and intrinsically noisy data. As such, finding trends in high dimensional spaces, and therefore rational search and purification strategies is difficult. Because of this, performing a grid search in a large crystallisation space is an inefficient strategy, making this a good system on which to apply Bayesian Exploration.

**Simulated Experiments**

In order to model the stochastic nature of polymorph crystallisation[19], a system with 4 possible polymorphs was initialised, where the likelihood of a single crystallisation event of a particular polymorph at a particular time point in an experiment followed a Poisson distribution (eq. 1), which is typical of crystallising systems[20]:

$$P_{(x;\lambda)} = \frac{\lambda^x e^{-\lambda}}{x!} \qquad (1)$$

where $x = 1$ for a single crystallisation, and $\lambda$ is the polymorph crystallisation rate, which is determined by equation (2)

$$\lambda = A\, e^{-\Delta G^*/kT} \qquad (2)$$

where A is a concentration independent pre-exponential factor based on molecular attachment rates and $\Delta G^*$ is the (concentration-dependent) nucleation free energy barrier given by

$$\Delta G^* = K \frac{\gamma^3 v_m^2}{ln^2(\sigma/\sigma^*)} \qquad (3)$$

where $K$ is a constant to represent temperature and the Boltzmann constant, $\gamma$ is the polymorph interfacial energy, $v_m$ is the polymorph molar volume, $\sigma^*$ is the polymorph solubility, and $\sigma$ is the concentration of the solution phase, which is determined by the number of moles, M, in solution divided by the solution volume v. The polymorph solubility at a point, $\sigma^*$ was simulated as the magnitude of a probability function of a Gaussian distribution, with randomly assigned mean and covariance, at that point. From this, a heatmap can be created of the phase likelihood at each point, (x) for each polymorph p by comparing the nucleation free energy barrier of this phase versus other polymorph phases:

$$P(p,x) \propto \frac{\Delta G^*_p}{\sum_k \Delta G^*_k} \qquad (4)$$

Simulated crystallisation conditions were generated using the same method as physical experiments, i.e. an initial set of solvents whose proportions were chosen at random from by a 3-dimensional Dirichlet distribution, as shown in Figure 3. Each simulated crystallisation experiment was initialised with a volume, $v_0$ and solution moles, $M_0$, and then proceeded in a series of timesteps, where at every point the volume of solvent was reduced by a fixed amount to simulate evaporation. This caused an increase in the solution concentration, which altered the crystallisation probability, $P_{(x,\lambda)}$, of each polymorph at that timestep and ratio of solvents. A crystallisation event for a particular polymorph was then determined by comparison of the crystallisation likelihood with a pseudorandom number between 0 and 1 taken from a uniform distribution. On a successful crystallisation event, a new polymorph crystal was initialised with a number of moles of compound (dependent on polymorph), $M_c$ and this value was subtracted from the number of moles of compound remaining in the solution phase. In each subsequent timestep, every crystal then grew in number of moles based on the equation:

$$M_c(t+1) = M_c(t)(1 + K_p c(t)) \qquad (5)$$

$M_c(t)$ is the number of moles of a crystal at timestep $t$, $K_p$ is the growth constant for polymorph p, and $c(t)$ is the solution concentration at timestep $t$.

The total number of timesteps was set so that the final solution volume would be 0, and the crystallisation parameters of each polymorph were such that complete conversion of solution compound moles to polymorph moles would be achieved in every experiment. These final values could then be compared against the theoretical proportions of each polymorph, which is determined by the ratio of initial likelihoods of each polymorph at timestep $t = 0$. Four methods of generation instantiation were simulated and compared: random, uncertainty minimisation, and estimated improvement 10 generations of 24 triplicate simulation crystallisations were performed. The simulations show that there is initially a large degree of uncertainty in the models (see SI Section 3 for more details).

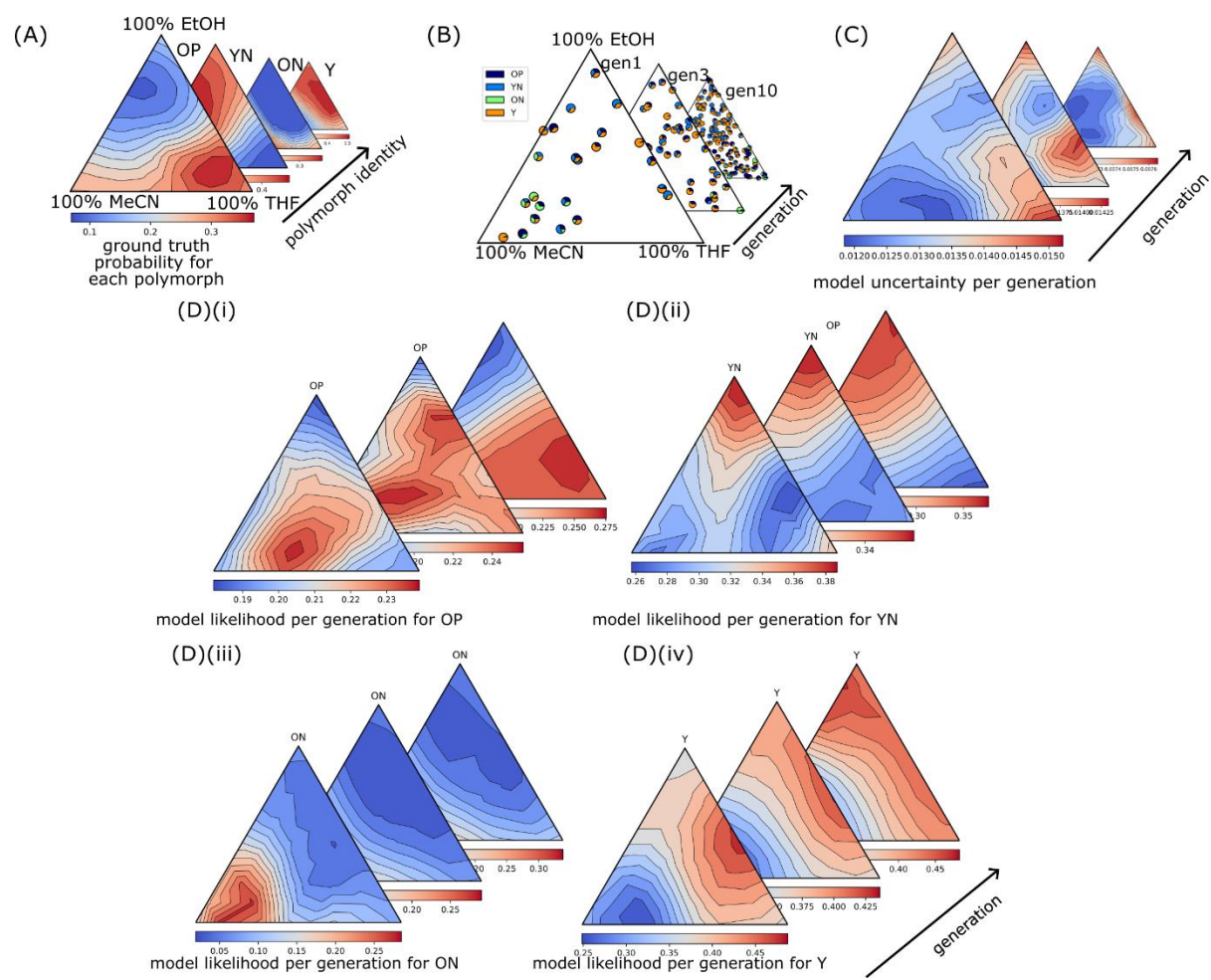

**Figure 3: Simulation of polymorph crystallization experiment.** A) Simulated ground truth likelihood heatmap for each polymorph in different solvent conditions (e.g. OP is much more likely to crystallize in THF than EtOH. MeCN is intermediate). B) Simulated experiments and results. The initial generation tests 24 solvent conditions at random. The relative amount of each polymorph formed in an experiment is represented by the fraction of each pie chart at that solvent point. Subsequent experiment conditions are determined by the model to either minimize overall model uncertainty, or maximize the likelihood of a particular polymorph. C) Simulated uncertainty heatmap of the model. Regions with few experiments and regions where nearby conditions give large differences in outcomes are more uncertain than highly explored regions with little difference between results. During exploration, the model uses uncertainty to decide subsequent experimental conditions. D) Simulated likelihood heatmap for each polymorph under different solvent conditions across generations. Over multiple generations, the likelihood approximates the ground truth for each polymorph shown in A).

**Crystal detection**

Every generation resulted in 72 images of crystals which needed to be classified and scored according to the relative amounts of each polymorph present in the sample. A general crystal classifier had been trained on data from the MARCO dataset, together with a set of samples of ROY crystals obtained from initial screening and was not retrained for the duration of the experiment. This could only classify one object, labelled as "crystal". For the identification and scoring of each polymorph, we began by assuming no knowledge *a priori* about polymorph appearance, abundance or identity. As such, images obtained in the first generation had to be manually labelled and the crystals undergo diffractometry where the polymorph identity after microscopic visual inspection was uncertain. Due to growth specific conditions, visually dissimilar crystals were occasionally found after XRD analysis to be the same polymorph, visually similar could be mislabelled as another polymorph, (e.g. some instances of R and OP). However, incorporating these disparate samples into the training data allowed the model to generalise and correctly classify similar samples, see Figure 4.

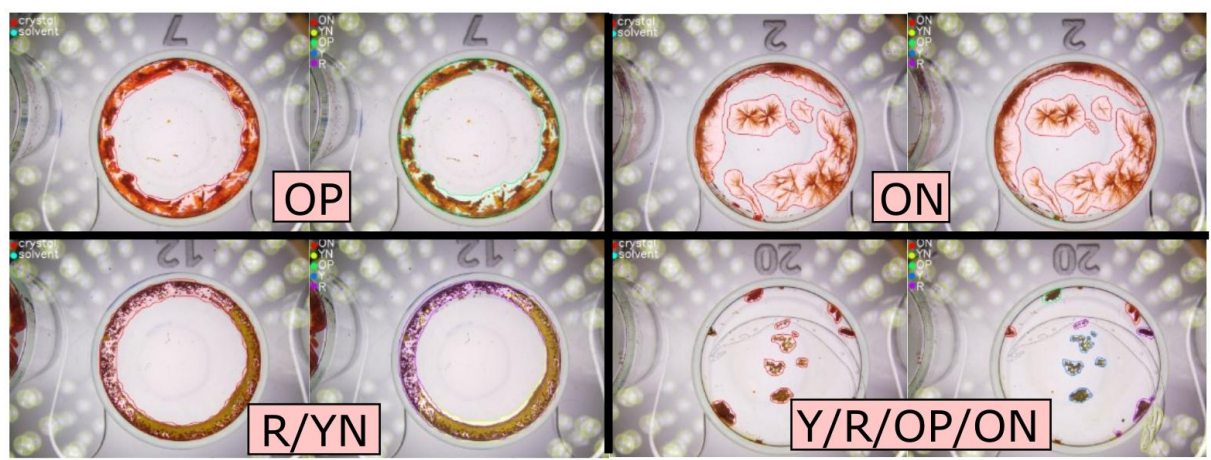

**Figure 4: Crystal polymorphs of ROY detected, segmented, and labelled using mask RCNN image detection using two detectors.** (1) identified the presence of crystalline material and (2) identified the presence of each type of crystal from the set [ON, YN, OP, Y, R] times. Top left- detection of OP only, top right- detection of ON only, bottom left detection of R and YN in the same vial. Bottom right Detection of Y, R, OP and ON in the same vial

The first generation was trained on 50 images of the polymorphs OP, ON, R, Y and YN. The classifier was retrained after generation 2 on a further 30 images in order to increase its generalisability and after generation 5 when polymorph ORP was confirmed by XRD. To assign a score for each polymorph in a reaction, the fractional area of each polymorph was calculated from the size of its pixel mask.

**Physical Experiments**

In the experiments, 7 generations of 24 triplicate crystallisations were performed. Each consisted of a 1ml solution of ROY in acetone (3.85 mmol L$^{-1}$) being mixed with a 1.5 ml mixture of four other solvents ($a$: ethanol, $b$: methanol, $c$: acetonitrile, and $d$: tetrahydrofuran) in different ratios, where $a + b + c + d = 1$. The values of each of the solvent variables were chosen at random for the first generation from a Dirichlet distribution. The image classifiers were those as described in the section on crystal detection, and the generation acquisition function used those as described in the simulation section, except that 4 solvent dimensions were investigated, and the simulated scores for each polymorph were replaced by the scores obtained from the image detection procedure outlined above.

Initially there was a large uncertainty in the models for each polymorph due to the large volume of unexplored space, but this decreased over successive generations. The veracity of the model can be checked when comparing the expected error between evaluated points and the model prediction at each of those points as seen in Figure 5a. However, the mean error difference between subsequent generations decreased over time, indicating that the model was converging on an overall solution.

This process was repeated for 7 generations, and the final surrogate models are shown in Figure 6. The first 5 generations saw only the first 5 initial polymorphs (ON, YN, OP, Y and R), however, in generation 6, a sample was classified as containing crystals but not as containing any known polymorph. The crystals in the sample were identified by x-ray crystallography as another polymorphic form of ROY, ORP, and the polymorph classifier was retrained to incorporate this, with the inclusion of an additional class.

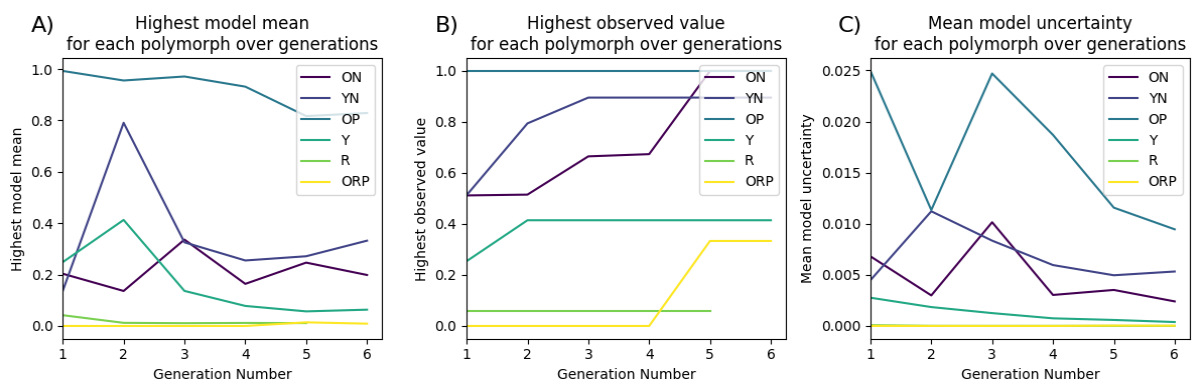

**Figure 5: Model state from experimental observations from generations of crystallization experiments.** (A) The highest value likelihood predicted by the model over each generation. These values are initially noisy, but are updated each generation and eventually converge as the model is exposed to more observations. (B) The maximum value obtained in a single experiment for each polymorph across each generation. (C) The model uncertainty for each polymorph which is updated each generation. Initially the results are noisy due to the probabilistic nature of crystallization, where sampling of the same data point can result in a different outcome. However, over subsequent generations, the model uncertainty decreases for each polymorph until the uncertainties stabilise.

However, since the number of instances of this polymorph was only 1, a generation was then created to confirm and establish the conditions required to produce ORP, as well as to gain additional training data for image detection. This was done by optimising for increased Expected Improvement (EI) of ORP over all the input parameters instead of decreased NIPV. Each of the Bayesian models was then updated to accommodate the new polymorph and the experiment was continued.

Over consecutive generations, the differences between the model prediction and actual experimental evaluations (error = |prediction-observation|) for each polymorph were seen to decrease as shown in Figure 4. The values in generation 0 were obtained with a prior assumption that all polymorphs occur with equal probability. The immediate drop in errors in the subsequent generation is due to incorporating the evaluations of the first set of results into the prior of the model, and the predictions from this being substantially greater than random. Over subsequent generations, the error decreases incrementally which reflects the model's continual improvement to the prior belief of the model.

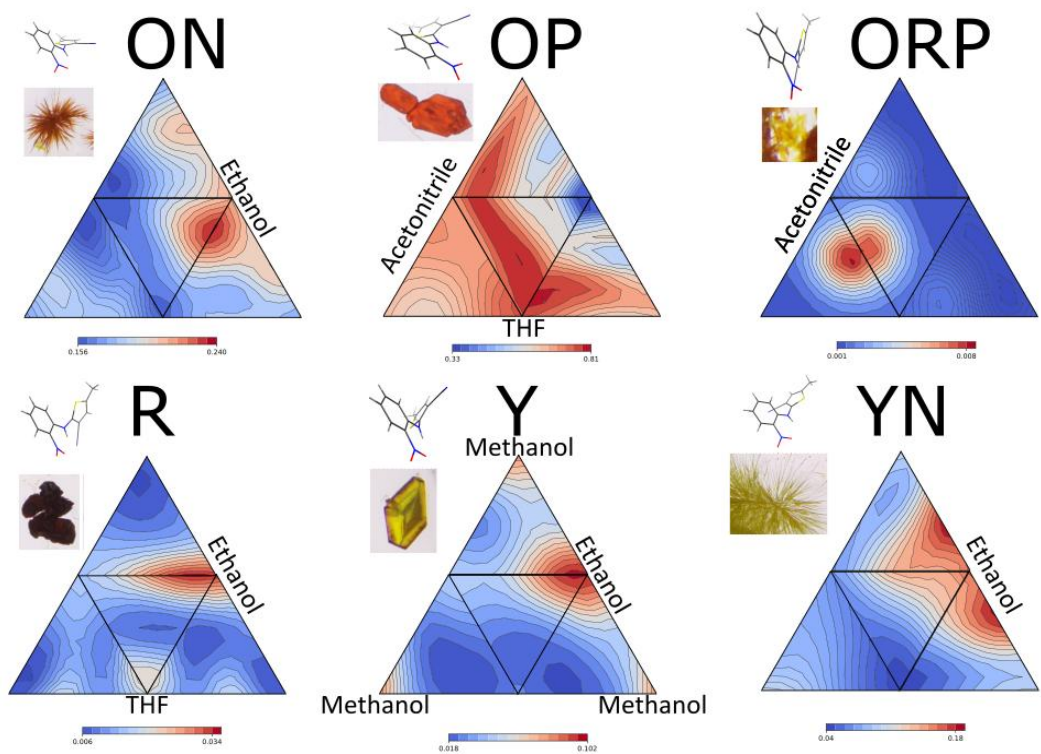

**Figure 6: Generation 7 experiments and observations.** Top left- stacked bar chart showing the ratios solvent composition for each reaction. Top right- compositional scatter plot of solvents used in each reaction. Vertices correspond to 100% of one solvent, opposite faces correspond to 0% of that solvent. Middle left-stacked bar chart showing the polymorph proportions observed for each reaction. Middle right- Polymorph proportions represented as pie charts located appropriately in a compositional scatter plot of solvents. Bottom- Plots showing four ternary heatmaps of the surrogate model where the sum of 3 components = 1, and the fourth is 0. Each heatmap corresponds to one surface of the tetrahedron above, and the whole can be viewed as a flattening out of the tetrahedron.

## Conclusions

We have shown that by using a combination of robotic automation, computer vision and artificial intelligence it is possible to run open-ended search and exploration loops for new polymorph forms in a complex polymorphic system. Our system was able to perform fully automated crystallisation cycles of a solvent space in triplicate to produce high-resolution images of multiple polymorphs of 5-methyl-2-[(2-nitrophenyl)amino]thiophene-3-carbonitrile (ROY). By employing two image classifiers trained on different features, first to determine the presence of crystals and second to classify the crystal form, we have shown that by using computer vision, different polymorphs can be automatically distinguished, with minimal human intervention. We have been able to discover the crystallisation conditions that lead to the rapid localisation and formation of a polymorph that had not previously been reported when using this combination of crystallisation solvents. As such, we believe this method is a viable strategy for the discovery of novel polymorphs for less well-explored compounds.

## Materials and Methods

**Chemical reagents.** All solvents used for crystallisation solutions and cleaning were HPLC grade from Sigma Aldrich. 5-methyl-2-[(2-nitrophenyl)amino]thiophene-3-carbonitrile (>97%) (ROY) was obtained from Tokyo Chemical Industry Ltd.

**Platform**. The platform was constructed in house from a range of 3D printed, laser-cut and commercially available components. Further details of the platform and an overview of the subassemblies can be seen in SI section 2 with a full bill of materials and component links can be found at the GitHub repository link in the same section. The software control of the platform for basic operations was written in Python 3. The software for image analysis was written in Python using Detectron2 and OpenCV packages.

**Acknowledgements**

We gratefully acknowledge financial support from the EPSRC (Grant Nos. EP/L023652/1, EP/R020914/1, EP/S030603/1, EP/R01308X/1, EP/S017046/1, and EP/S019472/1), the ERC (Project No. 670467 SMART-POM), the EC (Project No. 766975 MADONNA), and DARPA (Project Nos. W911NF-18- 2-0036, W911NF-17-1-0316, and HR001119S0003). We also

thank Yibin Jiang (Y.J.) for discussions regarding the implemented algorithms and Graham Keenan (G.K.) for software development.

**Author Contributions:**

**Corresponding Author**

Correspondence should be addressed to L.C.

**Data and materials availability**

The data used in this work are available upon request to the corresponding author via [Lee.Cronin@glasgow.ac.uk](Lee.Cronin@glasgow.ac.uk). The code used in this work is available at [https://github.com/croningp/Polymorph](https://github.com/croningp/Polymorph).